\pdfoutput=1

\documentclass[11pt]{article}

\usepackage[final]{acl}

\usepackage{times}
\usepackage{latexsym}
\usepackage{pifont}
\usepackage[T1]{fontenc}

\usepackage[utf8]{inputenc}

\usepackage{microtype}

\usepackage{inconsolata}

\usepackage{graphicx}
\usepackage{colortbl}
\usepackage{multirow}
\usepackage{booktabs}
\usepackage{amsmath}
\usepackage{amsfonts}
\usepackage{subfig}
\title{MoQAE: Mixed-Precision Quantization for Long-Context LLM Inference via Mixture of Quantization-Aware Experts}

\author{
 \textbf{Wei Tao$^{\spadesuit}$$^{\heartsuit}$},
 \textbf{Haocheng Lu$^{\spadesuit}$$^{\heartsuit}$},
 \textbf{Xiaoyang Qu$^{\heartsuit*}$},
 \textbf{Bin Zhang$^{\spadesuit}$$^{\heartsuit}$},
 \textbf{Kai Lu$^{\spadesuit}$\thanks{Xiaoyang Qu (email: quxiaoy@gmail.com) and Kai Lu (email: kailu@hust.edu.cn) are the corresponding authors.}},
 \textbf{Jiguang Wan$^{\spadesuit}$},
 \textbf{Jianzong Wang$^{\heartsuit}$}
\\
 $^{\spadesuit}$Huazhong University of Science and Technology,
\\
 $^{\heartsuit}$Ping An Technology (Shenzhen) Co., Ltd.
\\
 \small{
   \textbf{Correspondence:} \href{mailto:quxiaoy@gmail.com}{quxiaoy@gmail.com}, \href{mailto:kailu@hust.edu.cn}{kailu@hust.edu.cn}
 }
}

\begin{document}
\maketitle
\begin{abstract}
One of the primary challenges in optimizing large language models (LLMs) for long-context inference lies in the high memory consumption of the Key-Value (KV) cache. Existing approaches, such as quantization, have demonstrated promising results in reducing memory usage. However, current quantization methods cannot take both effectiveness and efficiency into account. In this paper, we propose MoQAE, a novel mixed-precision quantization method via mixture of quantization-aware experts. First, we view different quantization bit-width configurations as experts and use the traditional mixture of experts (MoE) method to select the optimal configuration. To avoid the inefficiency caused by inputting tokens one by one into the router in the traditional MoE method, we input the tokens into the router chunk by chunk. Second, we design a lightweight router-only fine-tuning process to train MoQAE with a comprehensive loss to learn the trade-off between model accuracy and memory usage.
Finally, we introduce a routing freezing (RF) and a routing sharing (RS) mechanism to further reduce the inference overhead. Extensive experiments on multiple benchmark datasets demonstrate that our method outperforms state-of-the-art KV cache quantization approaches in both efficiency and effectiveness.
\end{abstract}

\section{Introduction}

\begin{figure}
    \centering
    \includegraphics[width=0.48\textwidth]{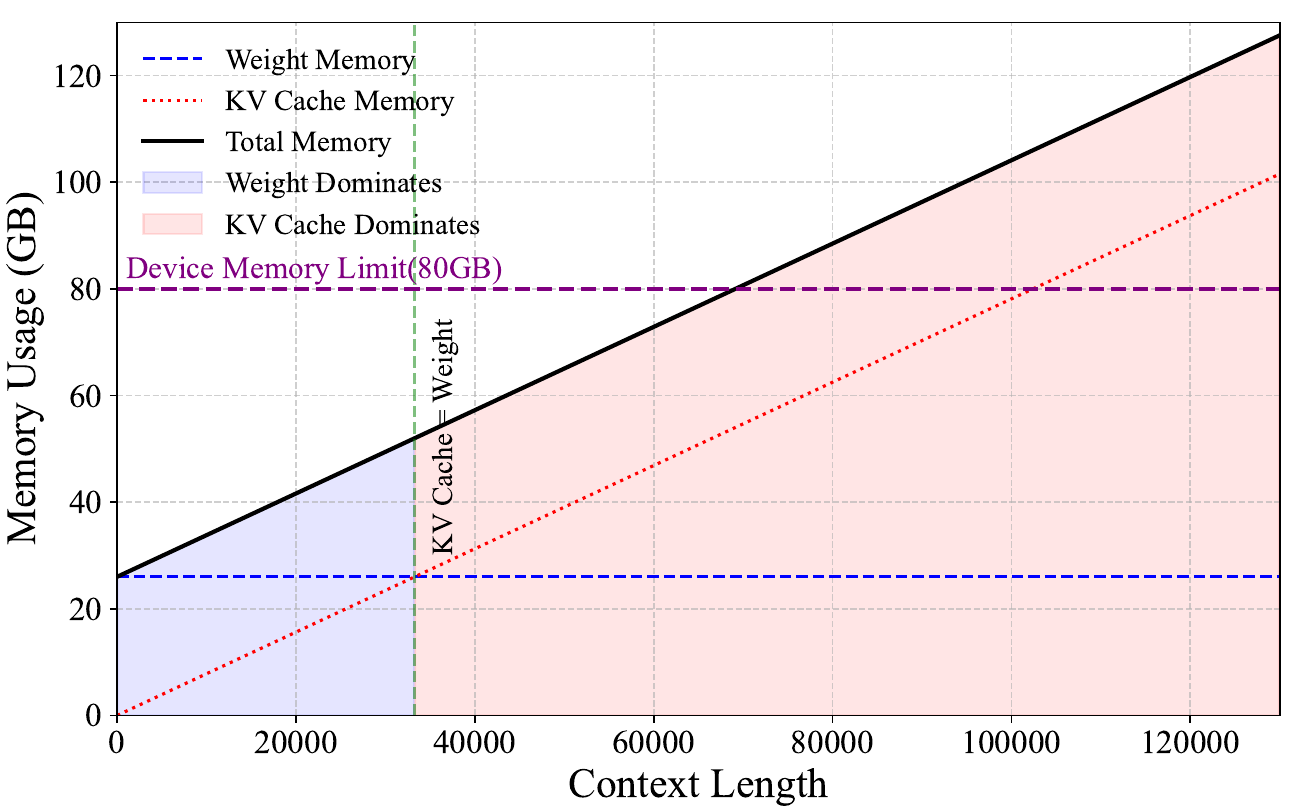}
    \caption{The composition of LLM inference memory under different context lengths on an NVIDIA A100 GPU with 80GB memory capacity.}
    \label{fig:background}
    \vspace{-5mm}
\end{figure}

In recent years, large language models (LLMs) have become a cornerstone in many fields, including natural language processing \cite{dubey2024llama}, computer vision \cite{lin2024video}, time series data \cite{tao2025madllm} and so on. As these models continue to evolve, the need to handle longer and more intricate texts has also grown significantly. Some complicatrd tasks often require models capable of handling extended contexts that span thousands of tokens. Although the newest LLM can handle up to 2 million input tokens \cite{team2024gemini}, the long-context inference still presents substantial challenges in memory consumption and computational efficiency. We have plotted the composition of the memory usage of the Llama2-13B model in relation to the context length in Figure \ref{fig:background} (The part beyond the device memory limit is our estimation). The memory occupied by the weights is fixed, while the memory occupied by the Key-Value (KV) cache is proportional to the context length. When the context length is small, the memory usage is still dominated by the weights. However, as the context length increases, it quickly shifts to being dominated by the memory usage of the KV cache. Ultimately, when the context length reaches 128k, the memory usage of the KV cache can reach 100GB, far beyond the memory capacity of most commodity GPUs. Obviously, during long-context inference, the main bottleneck in memory usage lies in the KV cache.
Furthermore, the frequent transfer of large KV caches between CPU and GPU memory for computation exacerbates the problem, leading to significant inference latency.

Researchers have proposed various methods to optimize LLMs for long-context inference, including pruning, knowledge distillation, and quantization. Among them, quantization is the easiest method to implement and can reduce memory consumption the most. Some researchers propose uniform quantizing models to low bit-width, which achieve great performance on memory reduction but can cause drastic accuracy degradation. Other researchers design mixed-precision quantization, which keeps the important tokens in high bit-width to maintain the model accuracy. However, these mixed-precision methods require complex and time-consuming quantization search processes to determine the bit-width configuration.

Inspired by MoICE \cite{lin2024mixture}, which employs the experts in the mixture of experts (MoE) module as the bases of rotary position embedding (RoPE), we leverage the advantages of the mixture of experts (MoE) approach's fast training and inference speed to propose MoQAE, a novel mixed-precision KV cache quantization method via mixture of quantization-aware experts. Our main innovation is to creatively use MoE technology to learn the quantization bit-width configuration. Specifically, our contributions consist of three components. (1) We treat each kind of quantization bit-width configuration as an expert (which is also the origin of the name "quantization-aware expert") and leverage the router in the MoE method to select the most suitable quantization bit-width. That is, 
we input a token into a router, which identifies the most suitable expert for that token. The quantization bit-width corresponding to that expert is the bit-width to which we need to quantize the token. We input tokens chunk-by-chunk instead of using the token-by-token manner in traditional MoE methods. (2) We design a lightweight fine-tuning process. Instead of training the entire LLM, we freeze the pre-trained LLM's parameters and perform minimal fine-tuning on the MoE routers using a calibration dataset. During fine-tuning, we introduce a comprehensive loss that balances model accuracy and memory usage. (3) We propose a routing-freezing (RF) and a routing sharing (RS) mechanism. The RF mechanism freezes the quantization strategy of initial chunks to keep model accuracy, while the RS mechanism allows the quantization strategy to be shared across different LLM blocks.

\section{Background}
\subsection{Preliminaries}
\textbf{LLM Inference.} Modern LLM architectures are predominantly based on a decoder-only structure, where inference is divided into two distinct stages: the prefill stage and the decoding stage. In the prefill stage, all input tokens are processed by the LLM to generate the first output token. Subsequently, during the decoding stage, a sequence comprising all input tokens and the tokens already generated is processed by the LLM to generate the next output token. This process repeats iteratively, with each newly generated token appended to the sequence for subsequent processing, until the entire output sequence is completed. A significant drawback of this approach is that, at each step, the key (K) and value (V) matrices corresponding to the input tokens and all previously generated tokens must be recomputed, leading to inefficiencies. To address this, modern LLMs utilize a KV cache, which stores the K and V matrices of both input and generated tokens, eliminating redundant computations and substantially reducing inference latency. However, when processing long input texts, the size of the KV cache grows dramatically, consuming a large amount of GPU memory and making model deployment infeasible on resource-constrained hardware. Moreover, the frequent transfer of the KV cache between CPU and GPU memory becomes more time-consuming as its size increases, turning the KV cache into a bottleneck for inference latency.

\textbf{Mixture of Experts.} MoE is a model architecture designed to divide computational tasks among multiple experts (sub-models) and dynamically select a subset of experts to process a given input using a routing mechanism. Recently, MoE architectures have been widely adopted in LLMs, such as Switch Transformer \cite{fedus2022switch} and GLaM \cite{du2022glam}. Traditionally, MoE treats each feed-forward network (FFN) layer in the LLM as an expert, and a router dynamically activates only a small subset of these FFN layers based on the input, while the inactive layers remain idle. This strategy has since been extended to self-attention layers as well \cite{zhang2022mixture}. Compared to dense models, MoE's sparse activation mechanism significantly reduces computational overhead while maintaining excellent scalability in parameter size. In this work, rather than viewing LLM layers as experts, we innovatively treat the quantization bit-width configurations of the KV cache in LLMs as experts and propose quantization-aware experts. 
\subsection{Related Works}
\textbf{KV Cache Optimization.}
Researchers have proposed various methods to optimize the KV cache in LLMs. Some \cite{zhang2023h2o, xiao2024efficient, han2024lm, liu2024scissorhands, ge2023model, pagliardini2023faster} have introduced pruning techniques to eliminate the KV cache of less important tokens. For example, Zhang et al. propose $\mathrm{H_2O}$ \cite{zhang2023h2o}, which removes tokens whose sum of vertical attention scores in the attention weight matrix is the lowest. StreamingLLM 
\cite{xiao2024efficient} proposes an ``attention sink" mechanism, and only keeps the initial tokens and the most recent tokens. Others \cite{song2024powerinfer, xue2024powerinfer, he2024fastdecode, kwon2023efficient, dao2022flashattention, yu2022orca, cai2024medusa, jin2023s} have focused on memory management strategies, addressing KV cache fragmentation from a system-level perspective. For instance, vLLM \cite{kwon2023efficient} constructs a page table that maps the continuous logical pages of the KV cache to non-contiguous physical memory pages, while also employing a copy-on-write mechanism to reduce memory usage. 
Jin et al. propose S3 \cite{jin2023s}, which predicts the output sequence length during inference and allocates KV cache memory space according to the prediction result, avoiding memory waste caused by over-allocating KV cache space. Additionally, quantization \cite{liu2024kivi, hooper2024kvquant, zhao2024atom, frantar2022gptq, yang2024no, kim2024squeezellm} has been explored as a promising approach to convert KV cache data from high-precision to low-precision formats, thereby saving memory. KIVI \cite{liu2024kivi} identifies the presence of many outlier channels in the key cache. Therefore, it proposes quantizing the key cache on a per-channel basis, while the value cache is quantized in the standard per-token manner. Atom \cite{zhao2024atom} applies asymmetric and 4-bit group quantization to the KV cache and performs dequantization before the KV cache computes with the query vector. Among these methods, quantization stands out as one of the most effective and straightforward solutions. However, traditional quantization often incurs significant performance degradation. In this paper, we propose a novel mixed-precision quantization method that achieves near-lossless model performance, addressing the limitations of existing techniques while optimizing KV cache memory usage.

\textbf{Mixed-Precision Quantization.} 
To mitigate the performance degradation caused by quantization, researchers have proposed mixed-precision quantization methods \cite{hooper2024kvquant, yang2024no, zhang2024q, kim2024squeezellm, lin2024awq, tao2025cocktail}. These approaches assign higher bit-widths to tokens of greater importance and lower bit-widths to less critical tokens, thereby maintaining model performance more effectively. In the beginning, researchers apply mixed precision quantization to the weights and activation values of LLM. For example, SqueezeLLM \cite{kim2024squeezellm} divides the weights of LLM into a dense matrix and a sparse matrix, and then uses INT8 quantization on the sparse matrix while keeping the precision of the dense matrix at FP16. AWQ \cite{lin2024awq} proposes an activation-aware weight quantization, which finds 1\% of salient weights through the distribution of activation values and reorders the weights to ensure hardware efficiency. Gradually, as the problems on the KV cache became increasingly prominent, mixed precision quantization has also been extended to the KV Cache. For example, MiKV \cite{yang2024no} uses the same method as $\mathrm{H_2O}$  to determine important tokens, but uses lower-bit quantization instead of evicting them. KVQuant \cite{hooper2024kvquant} retains high precision of the outlier value (value in large magnitude) in the KV cache during quantization, and designs a new data type nuqX to represent the KV cache after mixed precision quantization. However, most of these methods require a prohibitively long search time to determine the quantization bit-width. In this paper, we propose a novel mixed-precision quantization method via quantization-aware experts. This approach adopts the efficient routers in the MoE method to quickly and effectively learn the optimal quantization configuration for the KV cache.
\begin{figure*}
    \centering
    \includegraphics[width=\textwidth]{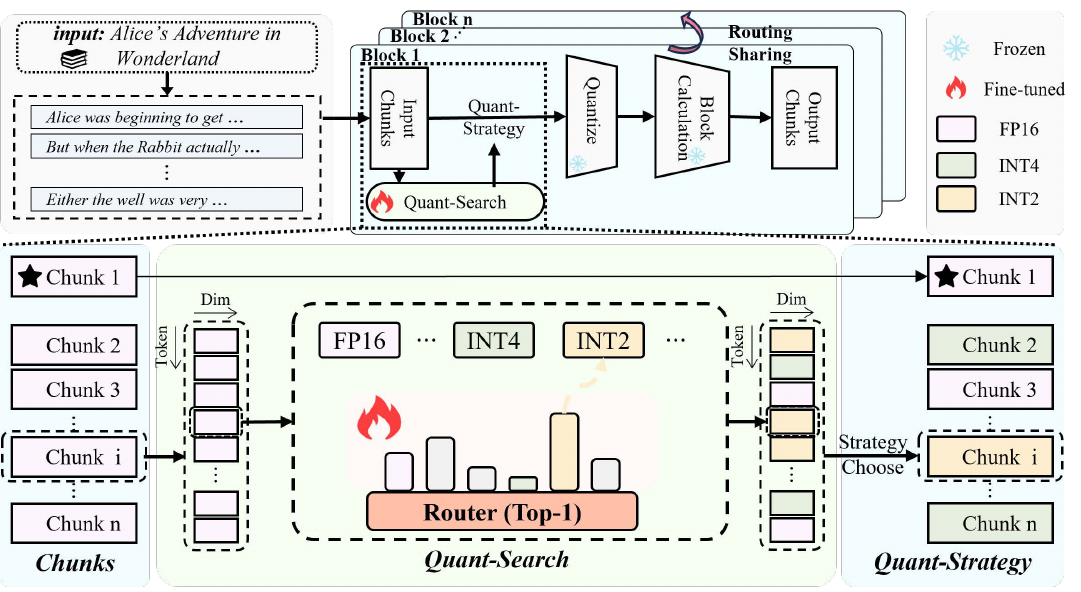}
    \caption{The overview of MoQAE. We use the router in MoE technology to learn the optimal quantization strategy. }
    \label{fig:overview}
    
\end{figure*}
\section{Method}
\subsection{Overview}
Figure \ref{fig:overview} shows the overview of MoQAE. The input text is first divided into several equal-length chunks, which are then processed by the LLM. In each block of the LLM, we use a quantization search module to determine the quantization strategy (i.e. quantization bit-width configuration) for the input chunks. Subsequently, these chunks are quantized using the bit-width configuration just determined, and proceeds with the formal calculation in the block (attention and feed-forward computations). Finally, the output chunk is passed to the next block, where the process is repeated. Notably, we apply a routing-freezing mechanism to the first chunk, preventing it from entering the router and fixing its bit-width to FP16. Additionally, we adopt a routing sharing mechanism between blocks, allowing different blocks to use the same quantization strategy.


\subsection{Quantization-Aware Experts}
In the quantization search module, we
introduce a router and several attention-aware experts. These experts represent different quantization bit-width configurations, such as FP16, INT4, INT2, and so on. The input text is divided into several equal-length chunks, and for the residual part that do not meet the chunk size, we directly retain their precision as FP16. Within each block of the LLM, the chunks are first passed into a router, where the router network is implemented using an MLP with the function:
\begin{equation}
    \mathcal{P} = f(CW_1\cdot CW_2)W_3
\end{equation}

Here, $C \in \mathbb{R}^{N \times D}$ is the input chunk, $f()$ is the activation function, $W_1, W_2 \in \mathbb{R}^{D \times M}$ and $W_3 \in \mathbb{R}^{D \times M}$ are weight matrices, where $D$ is the embedding dimension size within each attention head, $N$ is the chunk size, $M$ is the expert amount. The output $\mathcal{P} \in \mathbb{R}^{N\times M}$ reflects the probabilities of all the chunks about selecting which expert. 

For each token in the chunk, the expert with the highest selection probability is chosen as the selected expert for that token. 
Subsequently, we find out the expert that is selected the most times within the chunk and denote it as the quantization strategy for the entire chunk. The equation is as follows:
\begin{equation}
\begin{split}
    \mathcal{R} = \mathop{\arg\max}\limits_{1\leq k \leq M}\left( \sum_{i=1}^N \mathbb{I}\left( \mathop{\arg\max}\limits_{1\leq j \leq M} p_j^i = k \right) \right)
\end{split}
\end{equation}
Where $\mathcal{R} \in \{1,2,...,M\}$ is the quantization strategy, $p_j^i$ means the probability of selecting expert $j$ for chunk $i$, $\mathbb{I}\left(\right)$ operator means that the result is 1 if the condition is satisfied otherwise 0.
Finally, we integrate all the selected experts, generating the quantization strategy for all the chunks, and the input text will be quantized with this quantization strategy.

\subsection{Fine-Tuning Process}

To accelerate the training process, we design an efficient training method: freezing the parameters of the LLM itself and fine-tuning only the router's parameters. Additionally, our fine-tuning is conducted on a subset of the original dataset called the calibration dataset. 

\begin{figure}[ht]
    \centering
    \includegraphics[width=0.5\textwidth]{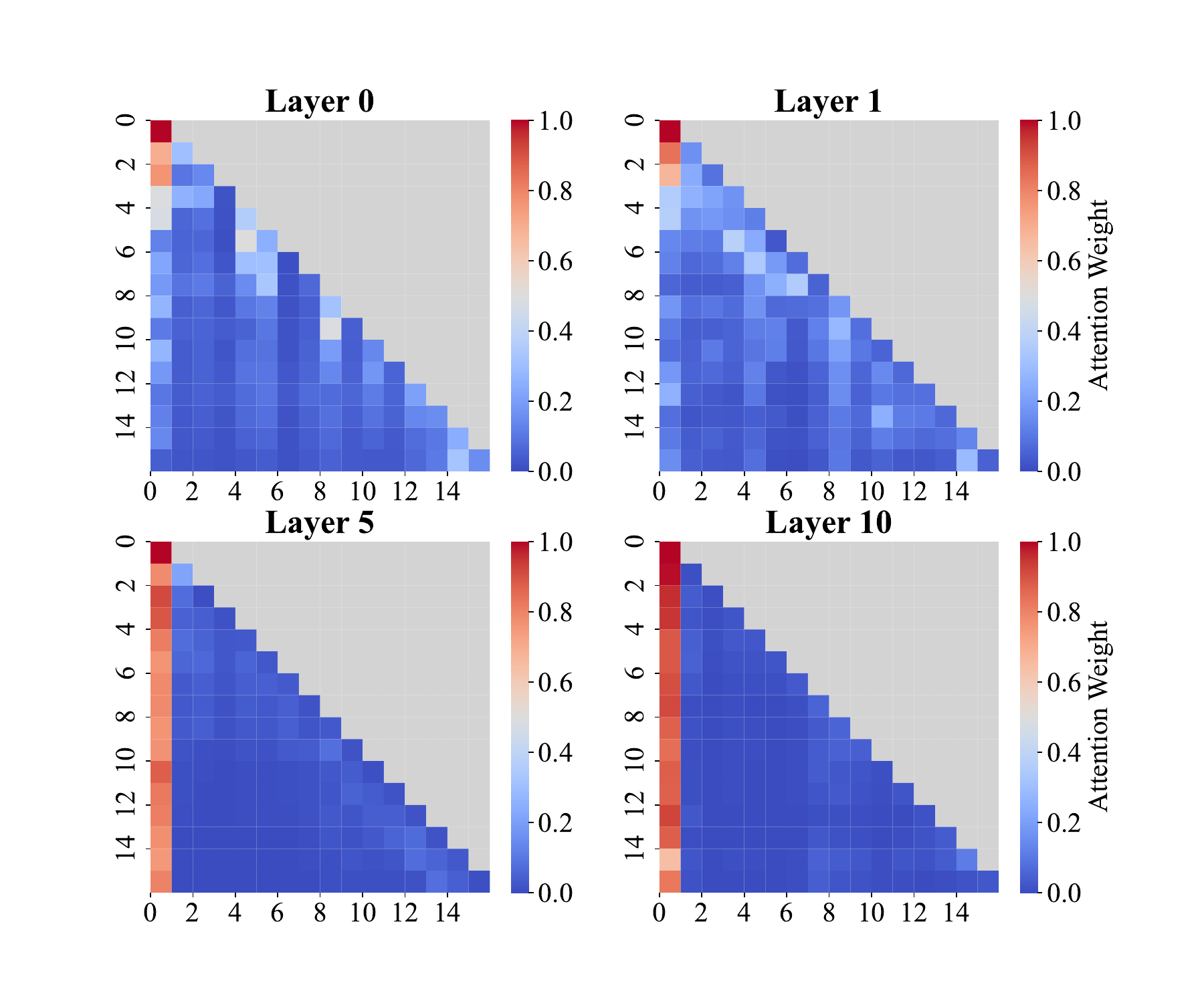}
    \caption{Attention weights of the first few tokens in different layers of Llama2-7b.}
    \label{fig:initial token}
    \vspace{-5mm}
\end{figure}
We further design a novel loss in the fine-tuning process. The goal of this loss is to achieve a trade-off between the accuracy of the LLM and memory usage during long-context inference. The design details of this loss are as follows:

On one hand, to optimize the model's accuracy, we incorporate the model's negative log-likelihood loss $L_{nll}$ as part of the final loss. However, we cannot directly apply $L_{nll}$ because it does not involve operators directly related to the router's weights, making it unable to train the router's weights. Therefore, we define a new loss called $L_{model}$, which is obtained by multiplying $L_{nll}$ by the mean value of the expert selection probabilities output by the router. To reflect the varying importance of different experts to the model's accuracy, we apply a penalty term to each component of this product. $L_{model}$ is ultimately computed as follows:
\begin{equation}
\begin{split}
    L_{model} =  
    \frac{1}{N}\sum_{i=1}^{N}\mathbb{I}\left( \mathop{\arg\max}\limits_{1\leq k \leq M}{p_k^i} = j \right)\cdot \frac{p_j^i\cdot L_{nll}}{B_j} \\
    \end{split} 
\end{equation}
where $p_k^i$ means the probability of selecting expert $k$ for chunk $i$, $1/B_j$ is the penalty term for expert $j$ and $B_j$ means the corresponding bit-width of expert $j$. We choose $1/B_j$ as the penalty term because data with lower bit-width leads to higher model loss.
\begin{figure}[ht]
    \centering
    \includegraphics[width=0.5\textwidth]{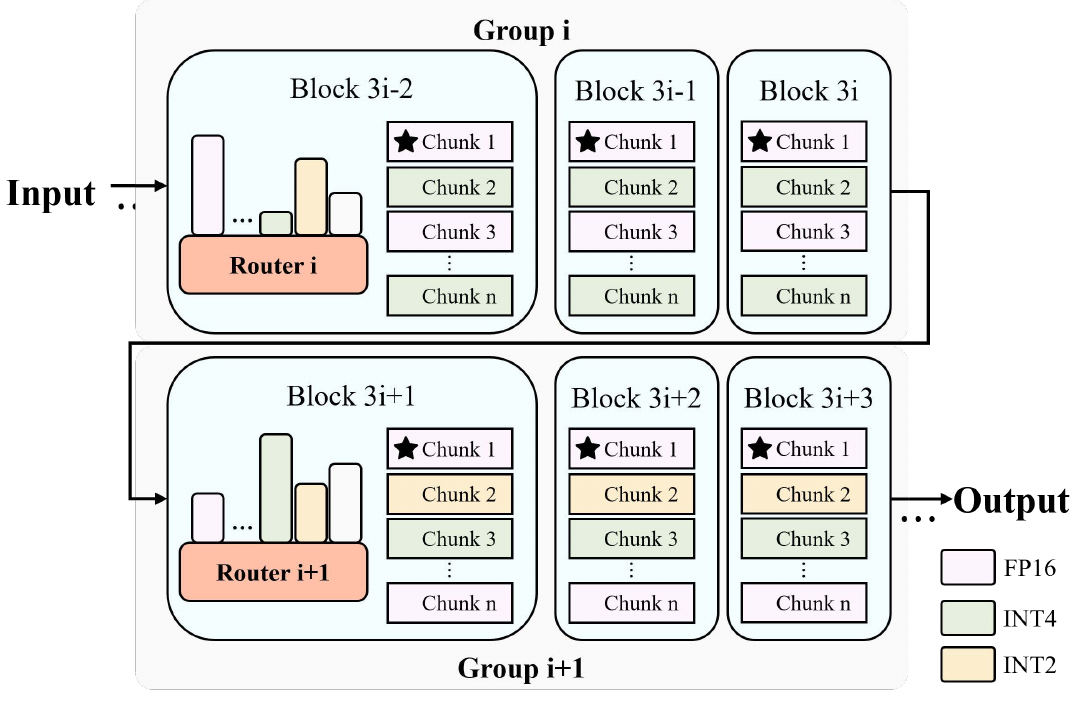}
    \caption{The routing sharing mechanism.}
    \label{fig:routing sharing}
    \vspace{-3mm}
    
\end{figure}

On the other hand, to ensure that our method also optimizes memory usage, we introduce the memory loss $L_{mem}$. The purpose of $L_{mem}$ is to encourage the router to preferentially select experts that represent lower bit-widths, thereby reducing the model's GPU memory usage. We also calculate $L_{mem}$ as the weighted sum of the mean value of the expert selection probabilities, but the penalty term is applied in an inverted manner:
\begin{equation}
\begin{split}
    L_{mem} = 
    \frac{1}{N}\sum_{i=1}^{N}\mathbb{I}\left( \mathop{\arg\max}\limits_{1\leq k\leq M}{p_k^i} = j \right)\cdot\frac{16p_j^i}{B_j}  \\
    \end{split} 
\end{equation}
Here we choose $\frac{16}{B_j}$ as the penalty term. This is because data with higher-bitwidth leads to more memory consumption.

Finally, our loss is defined as follows:
\begin{equation}
    L = \lambda L_{model} + (1-\lambda) L_{mem}
\end{equation}
where $\lambda$ is a pre-defined hyperparameter that controls the trade-off between model accuracy and memory usage. We will discuss the impact of $\lambda$ on model performance in Section \ref{subsec:ablation}. 
\subsection{Routing Freezing and Routing Sharing}

Previous researchers \cite{xiao2024efficient} have demonstrated that the token at the initial position of an LLM plays a crucial role in the model’s performance, significantly influencing its accuracy. In our research, we also explore this by conducting an experiment to investigate the attention weights of initial tokens of different layers within the LLM. As depicted in Figure \ref{fig:initial token}, we observe that the attention weights for tokens at the initial positions are relatively higher than those for tokens in subsequent positions (except for the first two layers). This finding strongly suggests that tokens at the beginning of the sequence are highly influential, playing a critical role in determining the model's output. These initial tokens seem to capture essential contextual information, which is then propagated through the rest of the sequence.

\begin{table*}[ht]
\centering
\caption{The perplexity of MoQAE and baseline methods on Wikitext2 dataset, lower is better. \textbf{AvB} means average bit-width. Most of the data is cited from CQ \cite{zhang2024kv}.}
\resizebox{\textwidth}{!}{
\begin{tabular}{lllccccc}
\toprule[1pt] 
\textbf{Bit Range} &
\textbf{Methods} &
\textbf{AvB} & \textbf{LLama-7B $\downarrow$} & \textbf{LLama-13B} $\downarrow$ & \textbf{LLama2-7B} $\downarrow$& \textbf{LLama2-13B} $\downarrow$& \textbf{Mistral-7B} $\downarrow$\\
\midrule
=16bits & FP16 & 16 & 5.68 & 5.09 & 5.11 & 4.57 & 5.07 \\
\midrule
\multirow{10}{*}{4$\sim$16bits} & INT4 \ding{172} & 4.00 & 7.40 & 6.82 & 7.31 & 6.59 & 5.91 \\
&INT4-gs128 \ding{172} & 4.16 & 7.16 & 6.67 & 6.87 & 6.20 & 5.76 \\
&NF4 \ding{173} & 4.00 & 7.27 & 6.74 & 7.09 & 6.45 & 5.85 \\
&NF4-gs128 \ding{173} & 4.16 & 7.16 & 6.66 & 6.86 & 6.20 & 5.77 \\
&KVQuant-4b \ding{174} & 4.00 & 7.13 & 6.65 & 6.70 & 6.11 & 5.75 \\
&KVQuant-4b-1\% \ding{174} & 4.32 & 7.09 & 6.62 & 6.65 & 6.06 & 5.72 \\
&CQ-2c8b \ding{175} & 4.00 & 7.11 & 6.64 & 6.67 & 6.09 & 5.74 \\
&Atom-4b-gs128 \ding{176} & 4.00 & 6.16 & 5.46 & 5.98 & 5.26 & 5.67 \\ 
&QoQ-4b \ding{177}& 4.00 & 5.93 & 5.28 & 5.88 & 5.32 & 5.62 \\
&QoQ-4b-gs128 \ding{177}& 4.00 & 5.89 & 5.25 & 5.89 & 5.24 & 5.66 \\
& AWQ \ding{178}& 4.00 & 6.33 & 5.59 & 6.51 & 5.43 & 6.24\\
& AWQ-gs128 \ding{178} & 4.00 & 5.93 & 5.36 & 5.92 & 5.27 & 5.66 \\
& MiKV \ding{179} & 5.50 & 6.25 & 5.58 & 5.89 & 5.33 & 5.78 \\
\rowcolor{orange!40}\cellcolor{white} & MoQAE-$\lambda$0.5 & 4.13 & \textbf{5.76}  & \textbf{5.15} & \textbf{5.22}  & \textbf{4.65}  & \textbf{5.14} \\

\midrule
\multirow{8}{*}{2$\sim$4bits}&INT2\ding{172} & 2.00 & 10892 & 100870 & 4708 & 4220 & 477 \\
&INT2-gs128\ding{172} & 2.14 & 43.49 & 56.25 & 113.49 & 97.04 & 50.73 \\
&NF2 \ding{173}& 2.00 & 2850.1 & 4680.3 & 13081.2 & 4175.6 & 1102.3 \\
&NF2-gs128 \ding{173}& 2.14 & 248.32 & 118.18 & 420.05 & 499.82 & 191.73 \\
&KVQuant-2b \ding{174}& 2.00 & 10.28 & 9.05 & 15.16 & 43.77 & 8.40 \\
&KVQuant-2b-1\% \ding{174}& 2.32 & 7.38 & 6.83 & 7.06 & 6.38 & 6.08 \\
&CQ-4c8b\ding{175} & 2.00 & 7.52 & 6.96 & 7.23 & 6.52 & 6.17 \\
&Atom-2b-gs128\ding{176} & 2.00 & 37.37 & 41.77 & - & - & - \\
\rowcolor{orange!40}\cellcolor{white} & MoQAE-$\lambda$0.3 & 3.50 & 8.17 & \textbf{6.44} & \textbf{6.26}  & 7.03  & \textbf{6.03} \\


\bottomrule[1pt]

  \end{tabular}}
\label{tab:performance}
\end{table*}
In response to these observations, we introduce a routing freezing mechanism to ensure that the critical tokens at the initial position are not compromised during the quantization process. Specifically, we prevent the first chunk of tokens from being passed into the router and restrict it to selecting the FP16 quantization configuration. This approach guarantees that the tokens at the start of the sequence are preserved with higher precision and are not quantized to lower bit-widths, thus protecting the model’s accuracy.

Additionally, we propose a routing sharing mechanism to optimize the inference process further. Our insight is inspired by CLA \cite{brandon2024reducing}, which demonstrates the feasibility of sharing key and value heads across different attention layers to reduce computational overhead. As illustrated in Figure \ref{fig:routing sharing}, in this mechanism, we partition the different blocks within the LLM into several groups. In each group, the other blocks share the quantization strategy of the first block. The routers in other blocks are also removed. By the routing sharing mechanism, we can effectively reduce the memory usage caused by too many routers and the latency caused by router computation in most of the blocks. Although sharing routing strategies between different blocks may lead to a slight loss in model accuracy (since the quantization strategy of the KV cache in one block may not be applicable to the next block), this loss is not very severe (We will prove it in Section \ref{subsec:ablation}). At the same time, the routing sharing mechanism can significantly reduce memory usage and computation latency. Therefore, we believe that this loss is acceptable. We also explore the impact of the group size on model performance in Section \ref{subsec:ablation}.

\section{Evaluation}
\subsection{Experimental Setup}

\begin{table*}[ht]
\caption{The performance of MoQAE and baseline methods on LongBench datasets, higher is better.}
\centering
\resizebox{\textwidth}{!}{
\begin{tabular}{lcccccccc}
\toprule[1pt]
\textbf{Method} & \textbf{Qasper $\uparrow$} & \textbf{QMSum} $\uparrow$ & \textbf{MultiNews} $\uparrow$ & \textbf{TREC} $\uparrow$& \textbf{TriviaQA} $\uparrow$& \textbf{SAMSum} $\uparrow$ & \textbf{LCC} $\uparrow$ & \textbf{RepoBench-P} $\uparrow$\\
\midrule
FP16  & 9.52 & 21.28 & 3.51 & \textbf{66.00} & 87.72 & 41.69 & 66.66 & 59.82 \\
KIVI-2b \ding{179}  & 9.26 & 20.53 & 0.97 & \textbf{66.00} & 87.42 & \textbf{42.61} & 66.22 & 59.67 \\
CQ-4c8b \ding{175} & 9.58 & 20.87 & 1.93 & \textbf{66.00} & 87.72 & 41.13 & \textbf{66.57} & 59.75 \\
MiKV \ding{179} & 9.14 & 20.63 & 0.85 & 65.88 & 87.21 & 41.44 & 66.18 & 59.55 \\
\rowcolor{orange!40} MoQAE & \textbf{9.79} & \textbf{21.23} & \textbf{3.47} & \textbf{66.00} & \textbf{87.89} & 41.37 & 66.53 & \textbf{59.94} \\
\bottomrule[1pt]
\label{tab:lb}
\end{tabular}}

\label{tab:example}

\end{table*}
\textbf{Benchmarks.} 
\begin{figure*}
    \centering
    \includegraphics[width=1\linewidth]{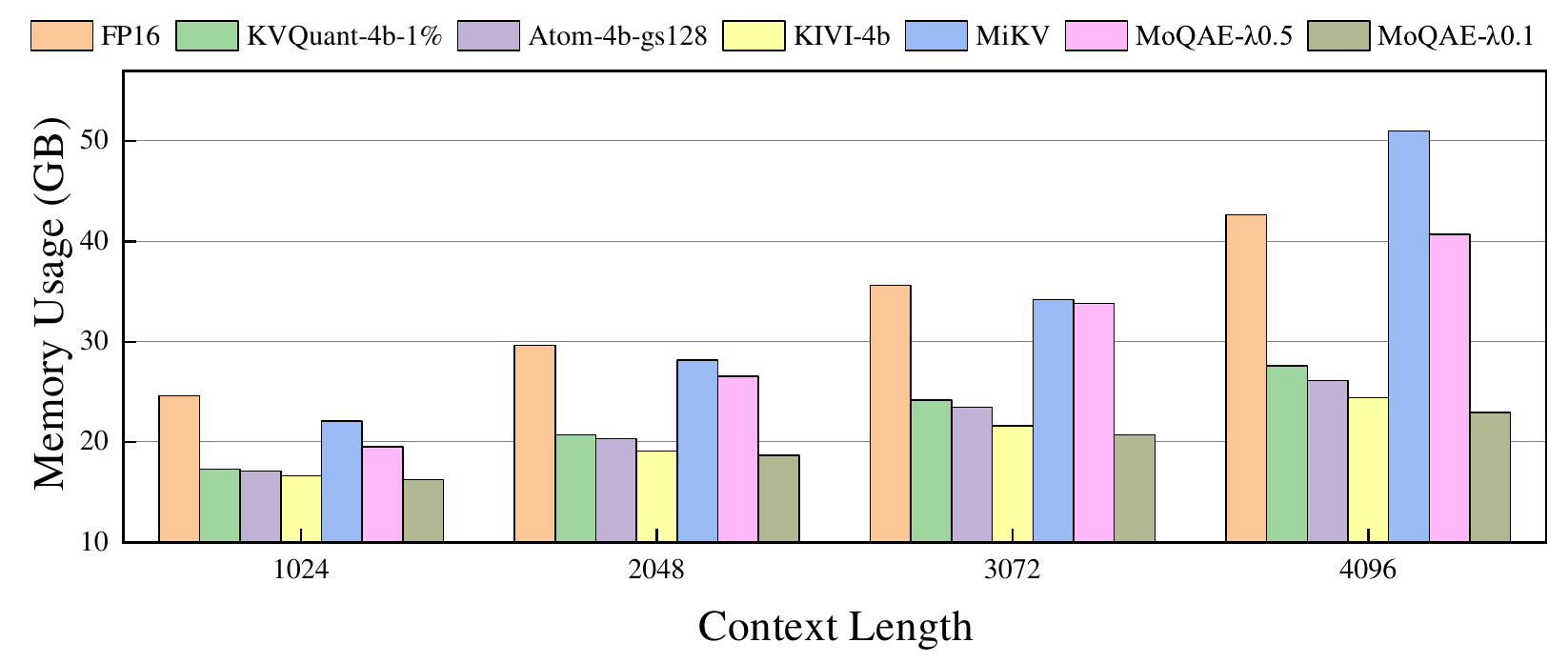}
    \caption{The memory usage of MoQAE and baseline methods under different context lengths.}
    \label{fig:context_memory}
\end{figure*}

We benchmark MoQAE on six widely-used open-source models: Llama-7B, Llama-13B\cite{touvron2023llama}, Llama2-7B, Llama2-13B \cite{touvron2023llama2}, Llama3-8B \cite{dubey2024llama}, and Mistral-7B \cite{jiang2023mistral}. To assess performance, we evaluate the perplexity of MoQAE on the WikiText2 \cite{merity2017pointer} dataset. We also adopt LongBench \cite{bai2024longbench} to further evaluate the long-context generation performance of our method and the baselines. We choose eight subsets from four different task types in LongBench as our practical datasets. They are single document QA task (Qasper), summarization task (QMSum, MultiNews), few-shot learning task (TREC, TriviQA, SAMSum), and code completion task (LCC, RepoBench-P). F1 score is used as the evaluation metric for Qasper and TriviaQA, while ROUGE score is used for QMSum, and MultiNews, and  similarity score is used for LCC and RepoBench-P. Only TREC uses classification score as the evaluation metric.
The maximum context length is 2048 for Llama, 4096 for Llama-2, Llama-3, and 8192 for Mistral, respectively.

\textbf{Baselines.} We compare MoQAE with the FP16 full precision model and nine other state-of-the-art KV cache quantization methods as the baselines: \ding{172} INT, which means uniform integer quantization. \ding{173} NF, which means NormalFloat quantization. \ding{174} KVQuant \cite{hooper2024kvquant}, which keeps outlier value in high bit-width. KVQuant-[$x$]b-1\% means 1\% of the tokens is kept as FP16 precision. \ding{175} CQ \cite{zhang2024kv}, which couples multiple key/value channels together to exploit their inter-dependency. CQ-[$x$]c[$y$]b means that each group has $x$ channels and there are $y$ bits in a quantized code for a group. \ding{176} Atom \cite{zhao2024atom}, which uses asymmetric uniform quantization with the granularity of attention head.  
\ding{177} QoQ \cite{lin2024qserve}, which scales queries and keys to decrease the loss caused by quantizing the outlier values in the key cache.
\ding{178} AWQ \cite{lin2024awq}, which applies uniform 4-bit quantization to the KV cache. \ding{179} MiKV \cite{yang2024no}, which employs mixed-precision quantization by computing the attention score sum of each token and quantizing those with low attention score sum to lower bit-width while keeping the rest at higher bit-width. \ding{180} KIVI \cite{liu2024kivi}, which uses per-channel quantization to the key cache and per-token quantization to the value cache. The
quantization bit-width for each token is assigned based on their
saliency. Among them, \ding{172}, \ding{173}, \ding{175}, \ding{176}, \ding{177}, \ding{178}, \ding{180} are uniform quantization; \ding{174}, \ding{179} are mixed-precision quantization. The suffix ``gs" in the method name indicates the group size, while other method names that do not contain "gs" means that those methods do not use group quantization.


\textbf{Implementation.} We conduct our experiments on an NVIDIA H20-NVLink GPU containing 96 GB of memory, along with a 25-core AMD EPYC 7T83 CPU and 100GB of RAM. Chunks size is set as 32, and $\lambda$ is set as 0.5. Group size in the routing sharing mechanism is set as 3. The router consists of a 2-layer MLP with a hidden dimension of expert amount. We use SiLU as the activation function and top-1 expert selection as the routing mechanism. The memory usage of the parameters of the router is about 1.6KB. As for training, we use 5\% of the full training set as the calibration dataset.
We use AdamW as the optimizer, with learning rate 3e-4 and batch size 8.

\begin{figure*}[ht]
    \centering
    \includegraphics[width=1\linewidth]{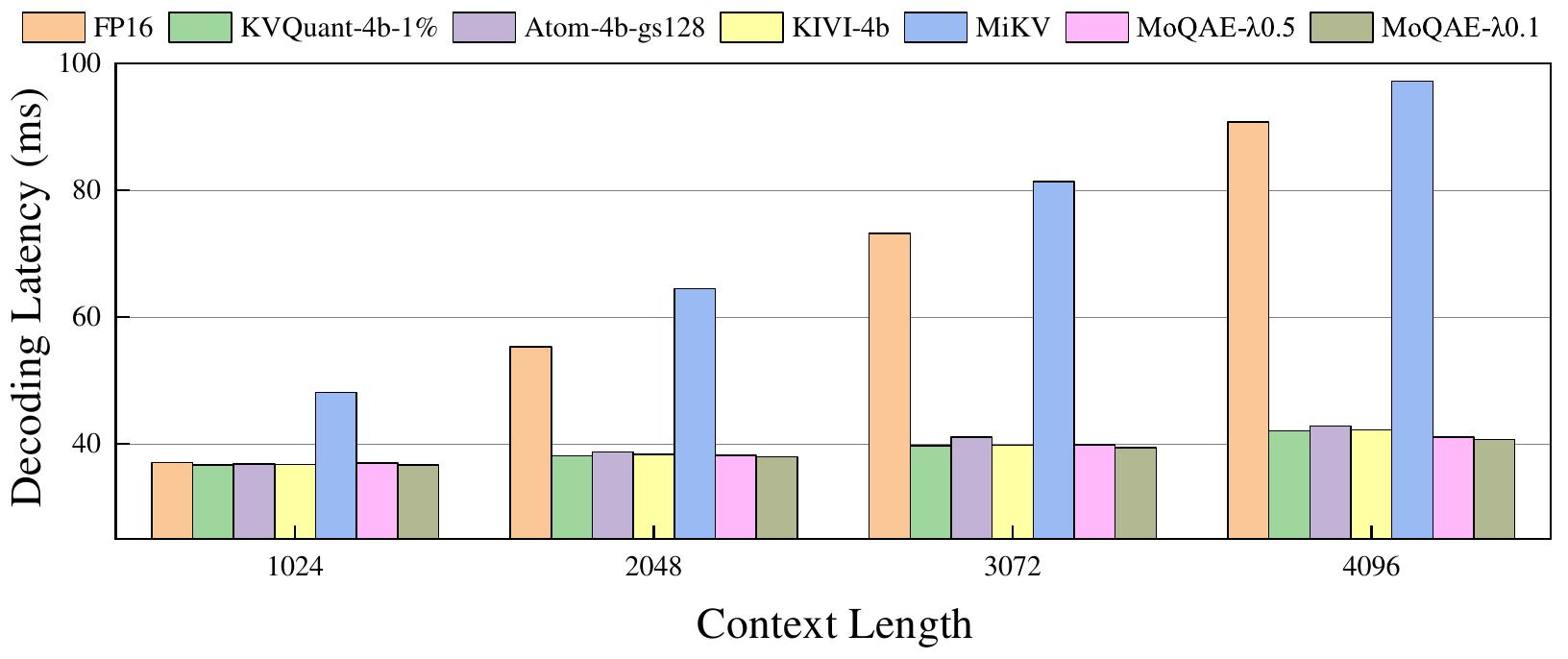}
    \caption{The decoding latency of MoQAE and baseline methods under different context lengths.}
    \label{fig:context_latency}
\end{figure*}

\begin{figure}[ht]
    \centering
    \includegraphics[width=0.5\textwidth]{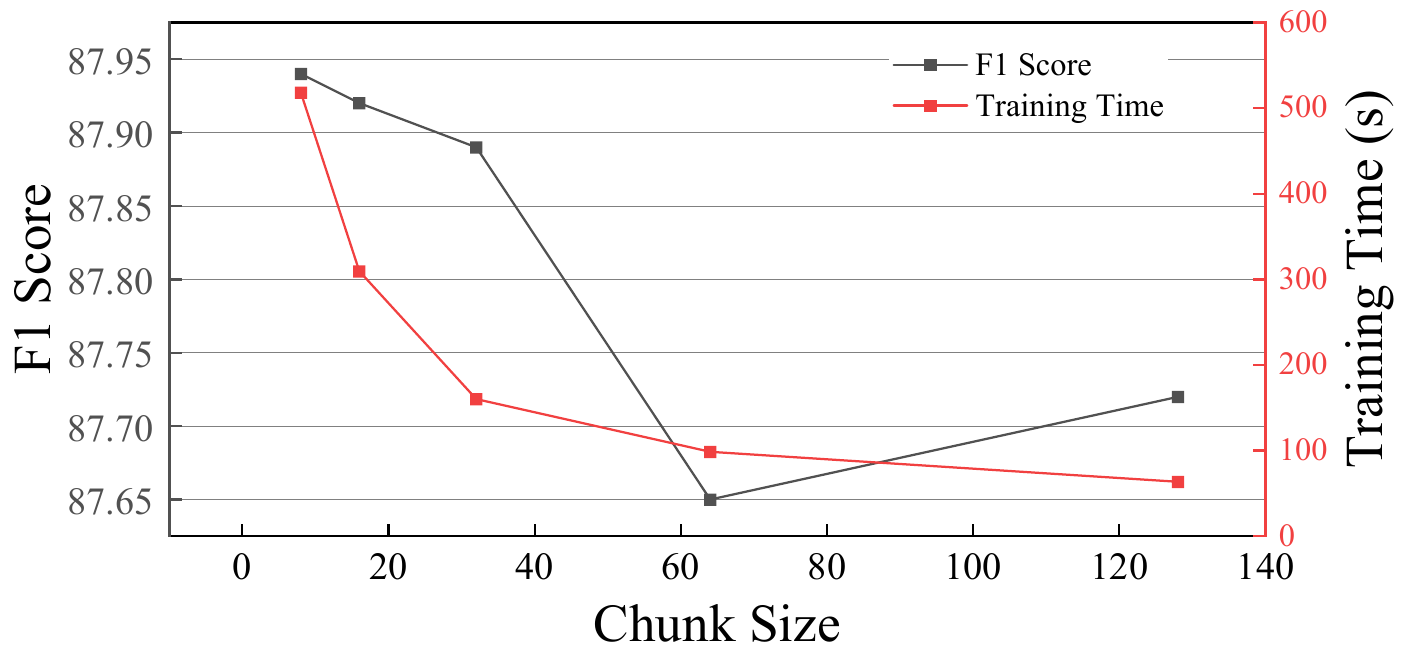}
    \caption{The impact of chunk size on model performance and training time.}
    \label{fig:chunk size}
   
\end{figure}

\subsection{Performance}
We first evaluate the perplexity on Wikitext2 dataset.
The results are shown in Table \ref{tab:performance}. We additionally test the case where $\lambda$ is 0.3. As can be seen from the table, simple quantization to extremely low bit-widths (2 bits) results in significant accuracy loss. Even with meticulously designed quantization methods, as the bit-width decreases, the model's accuracy rapidly declines. Compared to other methods, MoQAE is able to reduce the model's average bit-width to a relatively low level while maintaining model accuracy well. 
Among methods with 4-16 bits, MoQAE-$\lambda$0.5 achieves the least perplexity with similar average bit-width with baseline methods. The perplexity of MoQAE-$\lambda$0.5 is only 0.08 more than the FP16 models on average. MoQAE-$\lambda$0.3 also outperforms methods with 2-4bits on most models.

We also compare the performance of MoQAE and other
\begin{table}[ht]
    \centering
    \caption{The impact of chunk size on decoding latency.}
    \resizebox{0.5\textwidth}{!}{\begin{tabular}{cccccc}
    \toprule
        Chunk Size & 8 & 16 & 32 & 64 & 128 \\
        \midrule
        Decoding Latency/ms & 24.85 & 24.26 & 23.86 & 23.59 & 23.01 \\
        \bottomrule
    \end{tabular}}
    
    \label{tab:chunk_decoding}
\end{table}
methods on LongBench datasets. As shown in Table \ref{tab:lb}, MoQAE achieves the best performance on most of the datasets. The performance of MoQAE 
is only a little worse than baseline methods on SAMSum and LCC datasets.

Furthermore, we evaluate the memory usage and decoding latency of MoQAE and other methods under different context lengths with batch size 8. We test MoQAE under two kinds of $\lambda$. As shown in Figure \ref{fig:context_memory} and Figure \ref{fig:context_latency}, MoQAE-$\lambda$0.1 achieves the least memory usage and decoding latency over all the context lengths. 

Compared with the state-of-the-art (SOTA) quantization methods, MoQAE can 
reduce the memory usage by 0.79GB and reduce the decoding latency by 0.44ms, on average. The efficiency of MoQAE-$\lambda$0.5 is worse than MoQAE-$\lambda$0.1, but it still reduces the memory usage of FP16 model by 2.99GB on average and outperforms most of the baseline methods on decoding latency on decoding latency.

\subsection{Ablation Study}
\label{subsec:ablation}
We explore the impact of chunk size on model performance. The results are shown in Figure \ref{fig:chunk size} and Table \ref{tab:chunk_decoding}. As the chunk size increases, the training time decreases significantly and so does the decoding latency decreases. The model accuracy shows a trend of first decreasing and then increasing slightly. This is because when the chunk size becomes larger, some important token information will be wrapped in more unimportant token information within a chunk. Such a chunk may be misidentified as INT2 quantization by the router, resulting in the loss of important information. When the chunk size is large, since we fix the first chunk to FP16, more important information is saved, which slightly improves the model accuracy.

We further conduct ablation experiments on the hyperparameter $\lambda$. As shown in Table \ref{tab:lambda}, with the increase of $\lambda$, the model accuracy increases (The accuracy reaches the upper limit after $\lambda$ is greater than 0.5) while average bits and memory usage decreases. This result demonstrates that $\lambda$ can effectively balance model accuracy and memory usage.
\begin{table}[t]
    \centering
    \caption{The impact of $\lambda$ on model performance.}
    \resizebox{0.5\textwidth}{!}{
    \begin{tabular}{lccccc}
    \toprule[1pt]
     $\lambda$  &  0.1 & 0.3 & 0.5 & 0.7 & 0.9 \\
     \midrule
     F1 Score    & 87.32 & 87.64 & 87.89 & 87.91 & 87.92 \\
     Average Bits & 3.45 & 3.65 & 4.2 & 10.40 & 12.12\\
    Memory Usage/GB  & 14.01 & 14.04 & 15.95 & 15.33 & 15.88 \\
    \bottomrule[1pt]
    \end{tabular}}
    
    \label{tab:lambda}
\end{table}
\begin{table}[t]
    \centering
    \caption{The impact of our RF and RS mechanism. ``gs'' means group size in the RS mechanism.}
    \resizebox{0.5\textwidth}{!}{
    \begin{tabular}{lcc}
    \toprule[1pt]
        Method &  F1 Score & Decoding Latency/ms \\
        \midrule
        FP16 & 87.72 & 9.7 \\
        MoQAE w/o RF  & 87.88 & 20.6      \\
        MoQAE w/o RS  & 87.92 & 31.7  \\
        MoQAE (gs=2) &  87.92 &  25.7           \\
        MoQAE (gs=4) &  87.81     &   16.1
        \\
        MoQAE   & 87.89    & 20.7 \\
        \bottomrule[1pt]
    \end{tabular}}
    \label{tab:rs_ablation}
\end{table}
We also test the impact of routing freezing and routing sharing mechanisms. When routing freezing is removed from MoQAE, as can be seen from Table \ref{tab:rs_ablation}, both accuracy and inference latency are slightly reduced. This is because the first chunk of some blocks may change from the original fixed FP16 to other lower bit-widths. When routing sharing is removed, the decoding latency is significantly improved, while the accuracy is slightly increased. This is because after removing routing sharing, we need to perform more router calculations, but the calculated bit-width configuration will also be more accurate. At the same time, we test the impact of different group sizes in the routing sharing mechanism. It can be seen that as the group size increases, the decoding latency is significantly reduced, but the accuracy also slightly decreases.

\section{Conclusion}
In this paper, we introduce MoQAE, a novel mixed-precision quantization method based on mixture of quantization-aware experts. First, we treat different quantization bit-width configurations as  experts and apply the traditional MoE method to select the optimal configuration. To avoid the inefficiency of inputting tokens one by one in the conventional MoE method, we feed the tokens into the router chunk by chunk. Second, we propose a lightweight router-only fine-tuning process and design a novel loss that enables the model to learn the trade-off between model accuracy and memory usage. Finally, we introduce the RS and RF mechanisms, which further reduces the inference overhead caused by the routers. Extensive experiments on benchmark datasets show that our method outperforms SOTA mixed-precision quantization techniques in terms of both efficiency and effectiveness.

\section{Limitations}
Since our method introduces additional routers in LLM, the parameters of these routers will occupy a part of the memory, and the calculation of the router will also slow down the inference time of the model. Although we have adopted methods such as chunk input and routing sharing to optimize, these overheads still exist.

In addition, in order to ensure the accuracy of the attention calculation results, since softmax has high precision requirements when calculating the attention weight, we will dequantize the quantized key vector to FP16 and calculate it with the FP16 query vector. This dequantization operation will also cause additional delays.

\section{Acknowledgements}
This work was sponsored by the Key Research and Development Program of Guangdong Province under Grant No.2021B0101400003, the National Key Research and Development Program of China under Grant No.2023YFB4502701, the National Natural Science Foundation of China under Grant No.62172175, the China Postdoctoral Science Foundation under Grant No.2024M751011, the Postdoctor Project of Hubei Province under Grant No.2024HBBHCXA027.
    

\bibliography{custom}

\end{document}